\documentclass[10pt, conference, compsocconf]{IEEEtran}
\usepackage{graphicx}
\usepackage{multirow}
\usepackage{algpseudocode}
\usepackage{array}
\usepackage[cmex10]{amsmath}
\begin{document}
\title{ACO for Continuous Function Optimization: A Performance Analysis}
\author{
    \IEEEauthorblockN{Varun Kumar Ojha\IEEEauthorrefmark{1}, Ajith Abraham\IEEEauthorrefmark{1},V{\'{a}}clav Sn{\'{a}}{\v{s}}el\IEEEauthorrefmark{1}}
    \IEEEauthorblockA{\IEEEauthorrefmark{1}IT4Innovations, V{\v{S}}B Technical University of Ostrava, Ostrava, Czech Republic
    \\varun.kumar.ojha@vsb.cz,~ajith.abraham@ieee.org,~vaclav.snasel@vsb.cz}
}
\maketitle

\begin{abstract}
Performance of the metaheurisitc algorithms are often subjected to their parameter tuning. An appropriate tuning of the parameters drastically improves the performance of a metaheuristic. The Ant Colony Optimization (ACO), a population based metaheuristic algorithm inspired by the foraging behaviour of the ants, is no different. Fundamentally, the ACO depends on the construction of new solution variable by variable basis using a Gaussian sampling with a selected variable from an archive of solutions. The selection of variable for sampling plays a vital role in the performance of the ACO. Present study offers a performance analysis of the ACO based on the tuning of its underling parameters such as selection strategy, distance measure metric and pheromone evaporation rate. Performance analysis suggests that the Roulette Wheel Selection strategy enhance the performance of the ACO due to its ability to provide non uniformity and wider coverage of selection in terms of selection of a solution for sampling. On the other hand, distance measure metric, `Squared Euclidean' offers better performance than its competitor distance measure metrics. It is observed from the analysis that the ACO is sensitive towards the parameter evaporation rate. Therefore, a proper choice may improve its performance significantly. Finally, a comprehensive comparison between ACO and other metaheuristics suggested that the performance of the improvised ACO surpasses its counterparts.         
\end{abstract}


\IEEEpeerreviewmaketitle

\section{Introduction}
\label{sec:Intro}
Ant Colony Optimization (ACO) is a population based metaheuristic optimization algorithm inspired by the foraging behaviour of ants was initially proposed for solving the discreet optimization problems. Later, it was extended for the optimization of continuous optimization problems. In present study, we shall examine the strength and weakness of the classical ACO algorithm for function optimization. In present paper we shall follow the improvised version of the Ant Colony Optimization with an abbreviation ACO. The performance of the classical ACO is fundamentally dependent on the mechanism of the construction of new solutions variable by variable basis where an $n$ dimensional continuous optimization problem has $n$ variables. Therefore, the key to success of the ACO is in its construction of a new solutions. To construct a new solution, a variable from an available solutions archive is selected for a Gaussian sampling in order to obtained a new variable. Therefore, $n$ newly obtained variable construct a new solution. The other influencing parameter to the performance of the ACC is distance measure metric, which is used for computing the average distance between $i^{th}$ variable from a selected solution and the $i^{th}$ variables of the other solutions in an available solutions set (solution archive). In the distance computation, the parameter pheromone evaporation rate plays an important role. In present study we shall analyse the impact of the mentioned parameters on the performance of the ACO. We shall also analyse and compare the performance of the improvised ACO with the other classical metaheuristic algorithms.

We shall explore the minute details of the ACO for the optimization of the continuous optimization problems in section \ref{sec:caco}. A comprehensive performance analysis based on the underlying parameters of the ACO is provided in section \ref{sec:pa} followed by a discussion and conclusion in sections \ref{sec:con}.     

\section{Continuous Ant Colony Optimization (ACO)}
\label{sec:caco}
The foraging behaviour of the ants inspired the formation of a computational optimization technique, popularly known as Ant Colony Optimization. Deneubourg \textit{et al.} \cite{acoDeneubourg1990} illustrated that while searching for food, the ants, initially  randomly explore the area around their nest (colony). The ants secrete a chemical substance known as pheromone on the ground while searching for the food. The secreted pheromone becomes the means of communication between the ants. The quantity of pheromone secretion may depend on the quantity of the food source found by the ants. On successful search, ants returns to their nest with food sample. The pheromone trail left by the returning ants guides the other ants to reach to the food source. Deneubourg \textit{et al.} in their popular double bridge experiment have demonstrated that the ants always prefer to use the shortest path among the paths available between a nest and a food source. M. Dorigo and his team in early 90's \cite{acoDorigo1997ant,acoDorigoCaro1999} proposed Ant Colony Optimization (ACO) algorithm inspired by the foraging behaviour of the ants. Initially, ACO was limited to discrete optimization problems only \cite{acoDorigo1997ant,acoDorigoColorni1996,acoDorigo1999}. Later, it was extended to continuous optimization problems \cite{acoSochaDorigo2006}. Blum and Socha \cite{acoBlum2005training,acoKrzysztofSocha2007} proposed the continuous version of ACO for the training of neural network (NN). Continuous ACO is a population based metaheuristic algorithm which iteratively constructs solution. A complete sketch of the ACO is outlined in Figure \ref{alg:one}. Basically, the ACO has three phases namely, Pheromone representation, Ant based solution construction and Pheromone update. 

\subsubsection{Pheromone Representation}
Success of the ACO algorithm lies in its representation of artificial pheromone. The whole exercise of the ACO is devoted to maintain its artificial pheromone. The artificial pheromone represents a solution archive of a target problem. Socha and Dorigo \cite{SochaDorigo}, illustrated a typical representation of solution archive given in Figure \ref{fig:pheromone}. The solution archive shown in Figure \ref{fig:pheromone} contains $k$ number of solutions, each of which has $n$ number of decision variables. 
\begin{figure}[t] 
\centering
\begin{tabular}{c|c|c|c|c|c|c|c|c|c|c|}
\cline{2-7}\cline{9-9}\cline{11-11} $S_1$ & $s^1_1$ & $s^2_1$ & $\ldots$ & $s^i_1$ & $ \ldots $ & $s^n_1$ &  & $f(s_1)$ & & $ \omega_1 $\\ 
\cline{2-7}\cline{9-9}\cline{11-11} $S_2$ & $s^1_2$ & $s^2_2$ & $\ldots$ & $ s^i_2$ & $ \ldots $ & $s^n_2$ &  & $f(s_2)$ & & $ \omega_2 $\\  
\cline{2-7}\cline{9-9}\cline{11-11}        & $\vdots$ & $\vdots$ & $\ddots$ & $ \vdots$ & $\ddots$ & $\vdots$ &  & $\vdots$ & & $\vdots$ \\ 
\cline{2-7}\cline{9-9}\cline{11-11} $  S_j$ & $ s^1_j$ & $ s^2_j$ & $ \ldots$ & $ s^i_j$ & $  \ldots$ & $ s^n_j$ &  & $f(s_j)$ & & $ \omega_j $\\  
\cline{2-7}\cline{9-9}\cline{11-11}  & $\vdots$ & $\vdots$ & $\ddots$ & $ \vdots$ & $\ddots$ & $\vdots$ &  & $\vdots$ & & $\vdots$ \\
\cline{2-7}\cline{9-9}\cline{11-11}  $S_k$ & $s^1_k$ & $s^2_k$ & $\ldots$ & $ s^i_k$ & $ \ldots $ & $s^n_k$ &  & $f(s_k)$ & & $ \omega_k $\\ 
\cline{2-7}\cline{9-9}\cline{11-11} \multicolumn{1}{c}{}  & \multicolumn{1}{c}{$g^1$} & \multicolumn{1}{c}{$g^2$} & \multicolumn{1}{c}{} & \multicolumn{1}{c}{$g^i$} & \multicolumn{1}{c}{} & \multicolumn{1}{c}{$g^n$} & \multicolumn{1}{c}{} & \multicolumn{1}{c}{} & \multicolumn{1}{c}{} & \multicolumn{1}{c}{}\\ 
\end{tabular}  
\caption{A typical Solution Archive/Pheromone Table. In a sorted solution archive for a minimization problem, the function value associated with the solutions are $f(s_1) \ll f(s_2) \ll \ldots \ll f(s_k)$ therefore the weight associated with the solutions are $\omega_1 \gg \omega_2 \gg \ldots \gg \omega_k$. The weight indicated the best solution should have highest weight. For the construction of new solution $n$ Gaussian are sampled using a selected $\mu$ from the archive.}
\label{fig:pheromone}
\end{figure} 
In the case of a $n$ dimensional benchmark optimization problem, variables in solution $S_j$ indicates the variables of the optimization problem. Whereas, in the case of neural network (NN) training, a phenotype to genotype mapping is employed in order to represent NN as a vector of synaptic weights \cite{eannXinYao1993,eannAbraham2004}. Therefore, a solution $S_j$ in the archive represent a vector of synaptic weights or a solution vector. A solution vector is initialized or created using random value chosen from a search space defined as $\mathcal{S} \in \left[ min, max \right]$. I case of NN,  $min$ is defined as $x-\triangle x$ and $max$ is defined as $x+ \triangle x$, where $x$ is set to 0. In the case of discrete version of ACO, a discrete probability mass function is used whereas, in case its continuous version, a continuous probability density function is derived from pheromone table. The probability density function is used for the construction of $m$ new solutions. These $m$ new solutions are appended to initial $k$ solutions and then from $k + m$ solutions $m$ worst solutions are removed. Thus, the size of solution archive is maintained to $k$. 

\subsubsection{Ant Based Solutions Construction}
New solution is constructed variable by variable basis. The first step in the construction of new solution is to choose a solution from the set of solution archive based on its probability of selection. The probability of selection to the solutions in the archive is assigned using (\ref{eq:probWt}) or (\ref{eq:probFit}). The computation of the probability of selection a solution $j$ given in (\ref{eq:probWt}) is based on the rank of the solution in the archive whereas, the probability of selection given in (\ref{eq:probFit}) is based on the fitness value of the solution. For the construction of the $i^{th}$ ($i \in [1,n]$) variable of $l^{th}$ (index into new solution set i.e. $l \in [1,m]$) solution, $j^{th}$ ($j \in [1,k]$) solution from a solution archive is chosen based on its probability of selection as per (\ref{eq:probWt}) or (\ref{eq:probFit}). Several selection strategies may be adopted for the selection of $j^{th}$ solution from a solution archive. The method of selection may be based on the probability of selection assigned based on the fitness function value or the weight assigned to the solution based on their rank. Hence, the probability of choosing $j^{th}$ solution from the solution archive may be given as 
\begin{equation}
\label{eq:probWt}
p_j = \frac{\omega_j}{\sum\limits^k_{r=1} \omega_r },
\end{equation}
or
\begin{equation}
\label{eq:probFit}
p_j = \frac{f(s_j)}{\sum\limits^k_{r=1} f(s_r)}, 
\end{equation}
where $ \omega_j $ is a weight associated to the solution $j$ computed as 
\begin{equation}
\label{eq:wt}
\omega_j = \frac{1}{qk \sqrt{2 \pi}}e^{\frac{-(rank(j)-1)^2}{2q^2k^2} },
\end{equation}
where $q$ is a parameter of the algorithm. Since in (\ref{eq:probFit}) the smallest function value gets lowest probability, a further processing is required in order to assign highest probability to smallest function value. The mean of the Gaussian function is set to 1, so that the best solution acquire maximum weight. The $f(S_j)$ indicates the function value of the $j^{th}$ solution. In case of the optimization problems, function value computation is straight forward whereas, in the case of the neural network training, the fitness of the solution is assigned using the Root Mean Square Error (RMSE) induced on NN for the given input training pattern (a given training dataset) \cite{annHaykin1994}. The benchmark functions (including the RMSE computation of the benchmark dataset) are the minimization problems, therefore, the lower the value of a function higher the rank the solution in the solution archive. A detailed discussion on the selection methods is offered in section \ref{subsec:pa_sel}.
 
Once the $j^{th}$ solution is picked up, in the second step, it is required to perform a Gaussian sampling. A Gaussian distribution is given as  
\begin{equation}
\label{eq:gauss}
g(x,\mu,\sigma) = \frac{1}{ \sigma \sqrt{2 \pi}}e^{-\frac{(x-\mu)}{2\sigma^2} },
\end{equation}
where $\mu$ is $S^i_j$ and $\sigma$ is the average distance between the $i^{th}$ variable of the selected solution $S_j$ and the $i^{th}$ variable of the all other solutions in the archive. Various distance metrics adopted for computing the distance are comprehensively discussed in section \ref{subsec:pa_dist}. The Gaussian sampling parameter $\sigma$ may be expressed as
\begin{equation}
\label{eq:sigma_comn}
\sigma = \xi D_i,
\end{equation}
where the constant $\xi > 0$, is a parameter of the algorithm, known as pheromone evaporation rate (learning rate) and $D_i$ is the computed average distance between the selected solution and all the other solutions in the archive. For an example Manhattan distance $D_2$ may be given as 
\begin{equation}
\label{eq:d}
D_2 = \sum\limits^k_{r=1} \frac{\mid S^i_r - S^i_j \mid }{k -1}.
\end{equation}
\subsubsection{Pheromone Update}
In the final phase of the ACO, the $m$ number of newly constructed solutions are appended to initial $k$ solutions. The fitness of $k + m$ solutions are ordered in acceding sense. In the subsequent step, $m$ number of worst solutions are chopped out from $k + m$ solution. Thus, the size of solution archive is being maintained to $k$. The complete discussion about the ACO is summed up in the algorithm given in Figure \ref{alg:one}.
\begin{figure}[t]
\begin{algorithmic}[1]
\Procedure{ACO}{$k, n, \Delta x, f(.), m, \xi, \epsilon$} \Comment{$ k \rightarrow $ Archive size, $ n \rightarrow $ dimension, $f(.) \rightarrow$ objective function, $m \rightarrow$ \# of new solution , $\xi \rightarrow$ evaporation rate and $\epsilon \rightarrow $ stopping criteria.}
 \For{$i=1$ to $k$} 
   \For{$j=1$ to $n$}
		\State  $S_{ij} := rand(min, max)$
    \EndFor
    \State $f_i = function(S_i)$ \Comment Compute function value.
 \EndFor        
 \State $S := Sorting(S)$; \Comment Sorting in ascending order of $f$.
 
  \Repeat
    
    \For{$l=1$ to $m$}          
        \For{$i=1$ to $n$} 
          \State Choose a solution $S_{ji}$ according to probability of selection where $j$ $ \in $ $[1,k]$.
          \State $\mu_i  = S_{ji}$ and $\sigma_i$ as per (\ref{eq:sigma_comn})
          \State $S_{li}^{\prime} = \mathcal{N}(\mu_i,\sigma_i)$   
         \EndFor      
    	\State $f_l = function(S^{\prime}_l)$     
     \EndFor \Comment{$m$ new solution constructed} 
     
      \State  $S^{\prime \prime} = S + S^\prime$; \Comment $ |S|= k$, $|S^\prime| = m$ and $|S^{\prime \prime}| = k + m$, appending $m$ new solution to $k$ solutions.
        
     \State $S^{\prime \prime} = Sorting(S^{\prime \prime})$
	 
     \State $S := S^{\prime \prime} - S_m$; \Comment Truncate $m$ poor solutions.            
   \Until{Stopping criterion is satisfied}
   
  \Return $f_0$ \Comment Optimum function value.
\EndProcedure
\end{algorithmic}
\caption{Continuous Ant Colony Optimization (ACO)}
\label{alg:one}
\end{figure}

\section{Performance Evaluation}
\label{sec:pa}
The ACO algorithm mentioned in Figure \ref{alg:one}, is implemented using Java programming language. The performance of ACO algorithm is observed against the tuning of various underlying parameters. The parameters taken into consideration for performance analysis are, (i) Selection strategies used for the selection of solution for Gaussian sampling (ii) Distance metric  and (iii) pheromone evaporation rate. We shall test the algorithm over the benchmark function mentioned in the Table \ref{tab:testProblem}. The expression of the benchmark functions are as follows
\begin{equation}
\label{eq:Ackley}
 - a \exp\left(-b\sqrt{\frac{1}{d}\sum\limits_{i=1}^dx_i^2}\right)-\exp\left(\frac{1}{d}\sum\limits_{i=1}^d \cos(cx_i)\right)+ a + \exp(1),
\end{equation} 
\begin{equation}
\label{eq:Sphere}
 \sum\limits_{i=1}^d x_i^2,
\end{equation}
\begin{equation}
\label{eq:SumSquare}
 \sum\limits_{i=1}^d ix_i^2,
\end{equation}
\begin{equation}
\label{eq:Rosenbrock}
 \sum\limits_{i=1}^{d-1} [100(x_{i+1} - x_i^2)^2 + (x_i-1)^2],
\end{equation}
\begin{equation}
\label{eq:Rastring}
 10d + \sum\limits_{i=1}^{d} [x_i^2 - 10\cos(2\pi x_i)],
\end{equation}
\begin{equation}
\label{eq:Griewank}
 \sum\limits_{i=1}^d \frac{x_i^2}{4000} - \prod\limits_{i=1}^d\cos\left(\frac{x_i}{\sqrt{i}}\right)+1,
\end{equation} 
\begin{equation}
\label{eq:Zakarov}
 \sum\limits_{i=1}^d x_i^2 + \left(\sum\limits_{i=1}^d 0.5 i x_i\right)^2 + \left(\sum\limits_{i=1}^d 0.5 i x_i\right)^2, 
\end{equation}
\begin{equation}
\label{eq:DixonNprice}
 (x_1 -1)^2 \sum\limits_{i=1}^d i(2x_i^2 - x_i -1)^2,
\end{equation}
and
\begin{equation}
\label{eq:rmse}
\sqrt{\frac{1}{n}\sum\limits_{i=1}^n e_i^2} 
\end{equation}     
where in (\ref{eq:rmse}) $e_i = (\hat{y_i} - y_i)$ is the difference between the target value $y_i$ and predicted value $\hat{y_i}$ of a training dataset.
\begin{table}[b]
\centering
\caption{The benchmark optimization problems/functions considers for the experiment in order to evaluate the performance of ACO algorithms. The expressions and the range of the search space are mentioned.}
\label{tab:testProblem}
\fontsize{8}{10}
\selectfont
\setlength\extrarowheight{5pt}
\begin{tabular}{l|l|c|r|r|r}
\hline
	
\multicolumn{2}{c|}{Function}  & \multicolumn{1}{c|}{Expression} &  \multicolumn{1}{c|}{Dim.} & \multicolumn{1}{c|}{Range} & \multicolumn{1}{c}{$f(x^*)$} \\
\hline
F1 & Ackley			& as per (\ref{eq:Ackley})  & $d$  & -15,30      & 0.0 \\
F2 & Sphere	  		& as per (\ref{eq:Sphere}) & $d$  & -50,100     & 0.0 \\
F3 & Sum Square	    & as per (\ref{eq:SumSquare}) & $d$  & -10,10      & 0.0  \\
F4 & Dixon \& Price	& as per (\ref{eq:DixonNprice}) & $d$  & -10,10  & 0.0  \\
F5 & Rosenbrook		& as per (\ref{eq:Rosenbrock}) & $d$ & -5,10       & 0.0  \\
F6 & Rastring		& as per (\ref{eq:Rastring}) & $d$  & -5.12,5.12  & 0.0  \\
F7 & Griewank		& as per (\ref{eq:Griewank}) & $d$  & -600,600    & 0.0  \\
F8 & Zakarov		& as per (\ref{eq:Zakarov}) & $d$  & -10,10       & 0.0 \\
\hline
F9 & abolone(RMSE)	& \multirow{2}{*}{as per (\ref{eq:rmse})} & 90   & \multirow{2}{*}{-1.5,1.5} & 0.0  \\
F10 & baseball(RMSE)	&  & 170  & 	 & 0.0  \\
\hline
\end{tabular}
\end{table}

\subsection{Selection Method}
\label{subsec:pa_sel}
The selection of solution is critical to the performance of the ACO provided in Figure \ref{alg:one}. We shall analyse, how the selection strategies influence the performance the ACO. Several selection strategies may be adopted for the selection of solutions which have probability of selection assigned as per (\ref{eq:probWt}) (named as Weight)  or (\ref{eq:probFit}) (named as FitVal).  Now, we have Roulette Wheel Selection (RWS), Stochastic Universal Sampling (SUS) and Bernoulli Heterogeneous Selection (BHS) selection strategies available at hand. Therefore, each of these three selection strategy may be used to select an individual which has probability of selection assigned using either of the probability assignment method mentioned in (\ref{eq:probWt}) and (\ref{eq:probFit}). Therefore, we have six different strategies available at hand. The six selection strategies are namely, RWS(FitVal), RWS(Weight), SUS(FitVal), SUS(Weight), BHS(FitVal) and BHS(Weight).  

In Roulette Wheel Selection, the individuals occupy the segment of wheel. The slice (segment) of the wheel occupied by the individuals are proportional to the fitness of the individuals. A random number is generated and the individual whose segment spans the random number is selected. The process is repeated until the desired number of individuals is obtained. Each time an individual required to be selected, a random number $rand(0,1)$ is generated and tested against the roulette wheel. The test verifies that the random number $R$ fall in the span of which segment of roulette wheel. The individual corresponding to the segment to which the random number $R$ belongs to is selected. 

Unlike roulette wheel selection, the stochastic universal sampling uses a single random value to sample all of the solutions by choosing them at evenly spaced intervals. Initially, it is required to fix the number of candidates to be selected. Let, $k$ be the number of solution need to be selected. For the selection of first candidate, the random number $R$ in the case of SUS is generated in $[0, 1/k]$. For other candidates, random number $R_{i} := R_{i-1} + 1/k$ is obtained, where $i$ indicates individual to be selected. 

The Bernoulli Heterogeneous Selection (BHS) depend on the Bernoulli distribution \cite{Bernoulli} may be described as follows. For $k$ independent variables representing the function value of solutions, where $k$ indicates the number of solutions. Therefore, to select an individual, the BHS may act as follows. The Bernoulli distribution \cite{Bernoulli} is a discrete distribution having two possible outcomes labelled by $sel = 0$ and $sel = 1$ in which $sel = 1$ ("success") occurs with probability $pr$ and $sel=0$ ("failure") occurs with probability $qr = 1-pr$, where $0<pr<1$ . Therefore, it has probability density function 
\begin{equation}
Pr(sel) =
\begin{cases}
1-pr & \text{for $sel = 0$}\\
pr & \text{for $sel = 1$.}
\end{cases}
\end{equation}

Accordingly, the outline of Bernoulli Heterogeneous Selection algorithm proposed in Figure \ref{alg:bhs}.
\begin{figure}[!h]
\begin{algorithmic}[1]
\Procedure{BHS}{$P$} \Comment $P \rightarrow$ vector containing probability of selection of the individuals in a population $k$
 \For{$j$ = $1$ to $k$}
   \State $R := rand(0,1)$ \Comment random value
   \If{(select($R, p_j$))}
   
      \Return Solution $j$ is selected
   \EndIf
 \EndFor
\EndProcedure 
\Procedure{select}{$R,p_j$} 
   \If{($R$ $<$ $p_j$)}
     
      \Return true
   \Else   
   
      \Return false    
   \EndIf
\EndProcedure
\end{algorithmic}
\caption{Bernoulli Heterogeneous Selection (BHS)}
\label{alg:bhs}
\end{figure}

The parameters setting for the performance evaluation of ACO based on selection strategy is as follows, the solution archive  $k = 10$, $n = 30$, $m = 10$, $\xi = 0.5$, $ \epsilon = $ 1000 iterations and distance metric chosen is D2 (Manhattan). The experimental results of the various selection strategy is provided in Table \ref{tab:sel_sse} where the values are the mean of the functions F1 to F10 listed in Table \ref{tab:testProblem} where the each function F$_i$ has and its value computed over an average of 20 trials. In other words, the each function values are the average over 20 distinct trials/instances and the final value given in Table \ref{tab:testProblem} is the average of each function. 
\begin{table}
\caption{An experimental result for the performance evaluation of selection strategies. The results indicated the superiority of RWS method over other selection strategies}
\label{tab:sel_sse}
\centering
\setlength\extrarowheight{2pt}
\begin{tabular}{p{2.5cm}|c c c c}
\hline
Selection probability assignment & \multicolumn{4}{c}{Selection Method}\\
\cline{2-5}
                          & \multicolumn{1}{c}{RWS}
                          & \multicolumn{1}{c}{SUS}
                          & \multicolumn{1}{c}{BHS}
                          & \multicolumn{1}{c}{Rank 1}\\
\hline
function fitness value based  & 28.202 & 101.252 & 34.931 & \multirow{2}{*}{41.725}\\                      

weight computed based on rank  & 35.105 & 95.216 & 36.503 &\\ 
\hline
\end{tabular}
\end{table}
Examining Table \ref{tab:sel_sse}, it may be observed that the RWS selection strategy outperform the other selection strategies.
The RWS selection with the probability of selection computed based on function value yields best result among the mentioned selection strategies. The performance of BHS selection strategy is competitive to RWS selection. The result in table indicate the   perform SUS is worst among the all mentioned selection methods. From Table \ref{tab:sel_sse}, it is interesting to note that performance over probability of selection based on function value performs better than its counterpart with an exceptional being SUS case. 
   
\begin{figure}[!t]
\centering
\includegraphics[width=3.2in,]{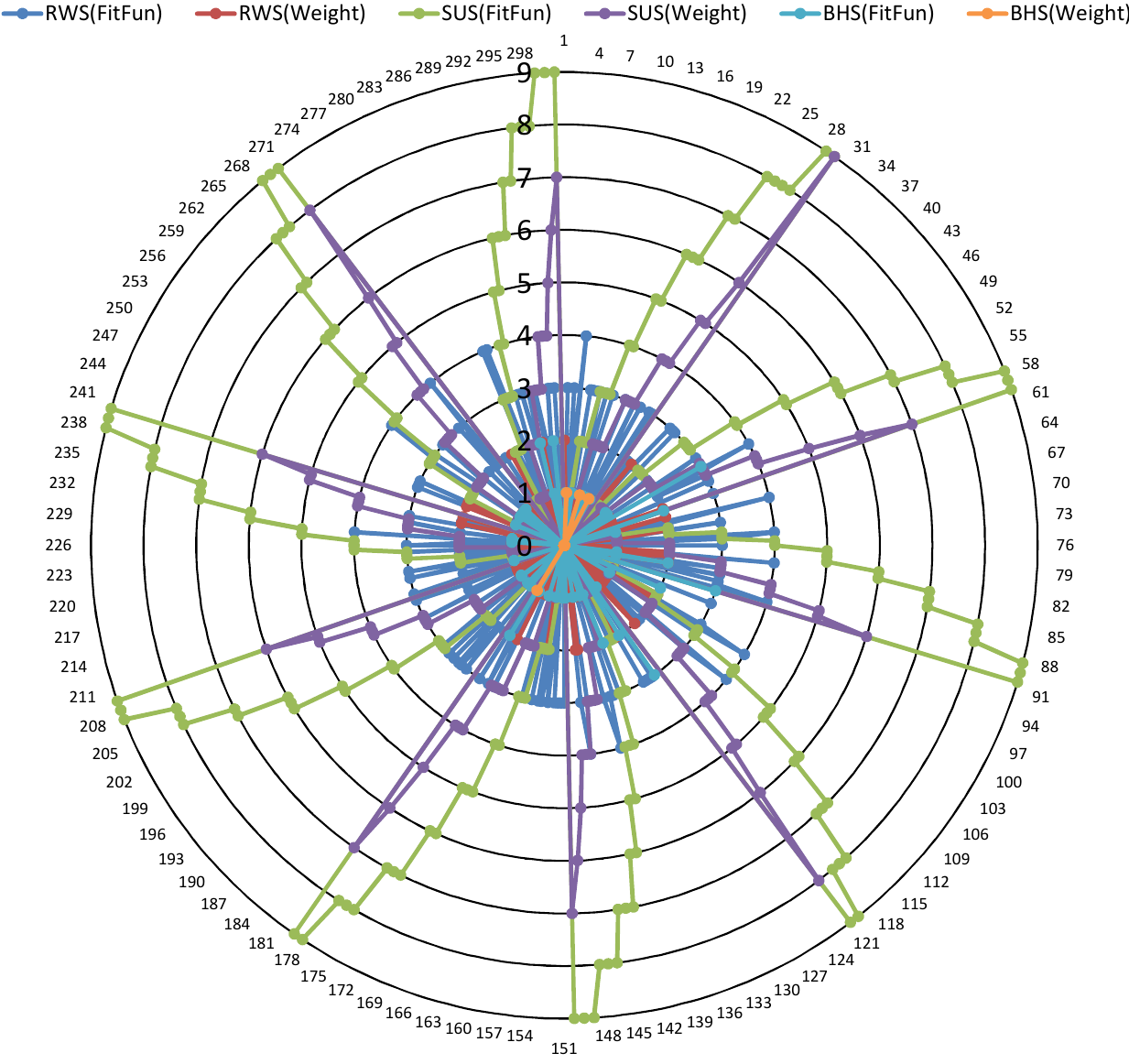}
\caption{A combined representation of the selection strategies. The center represent an individual with rank 1 and outermost circle represents an individual with rank 10.}
\label{fig:sel}
\end{figure}

To investigate the differences in the performance of the selection strategies indicated above, three hundred selections by each of the selection methods is plotted graphically in Figure \ref{fig:sel}. As per the algorithm outlined in Figure \ref{alg:one}, to construct a single solution, variable by variable, we need to select a solution from the solution archive of size $k$ individuals $n$ number of times, where $n$ is the number of variables in a solution. Hence, to construct $m$ number of solutions each of which having $n$ variables $m \times n$ selection is required. Figure \ref{fig:sel} represents the construction of 10 new solution from a solution archive of size 10 and each solution in the archive is having 30 variables representing the dimension of a function F1. Therefore, 10 $\times$ 30 = 300 selection are made at one iteration. Figure \ref{fig:sel} illustrate the mapping of selection made in a single iteration of the algorithm ACO used for optimization of function F1. In figure \ref{fig:sel}, ten concentric circles represent the solutions in the solution archive. The center of the circle (marked 0) indicate the solution with rank 1 (highest) while the subsequent outer concentric circle indicates the representation of increasing rank of solution. Therefore, the center indicated the best solution whereas the outermost concentric circle indicate the worst solution. Hence from the center to the outermost concentric circle each circle in increasing diameter represents $2^{nd}$, $3^{rd}$, $4^{th}$, $5^{th}$, $6^{th}$, $7^{th}$, $8^{th}$, $9^{th}$ and $10^{th}$ ranked solution respectively.

\begin{figure}[!t]
\centering
\includegraphics[width=1.5in, height = 1.5in]{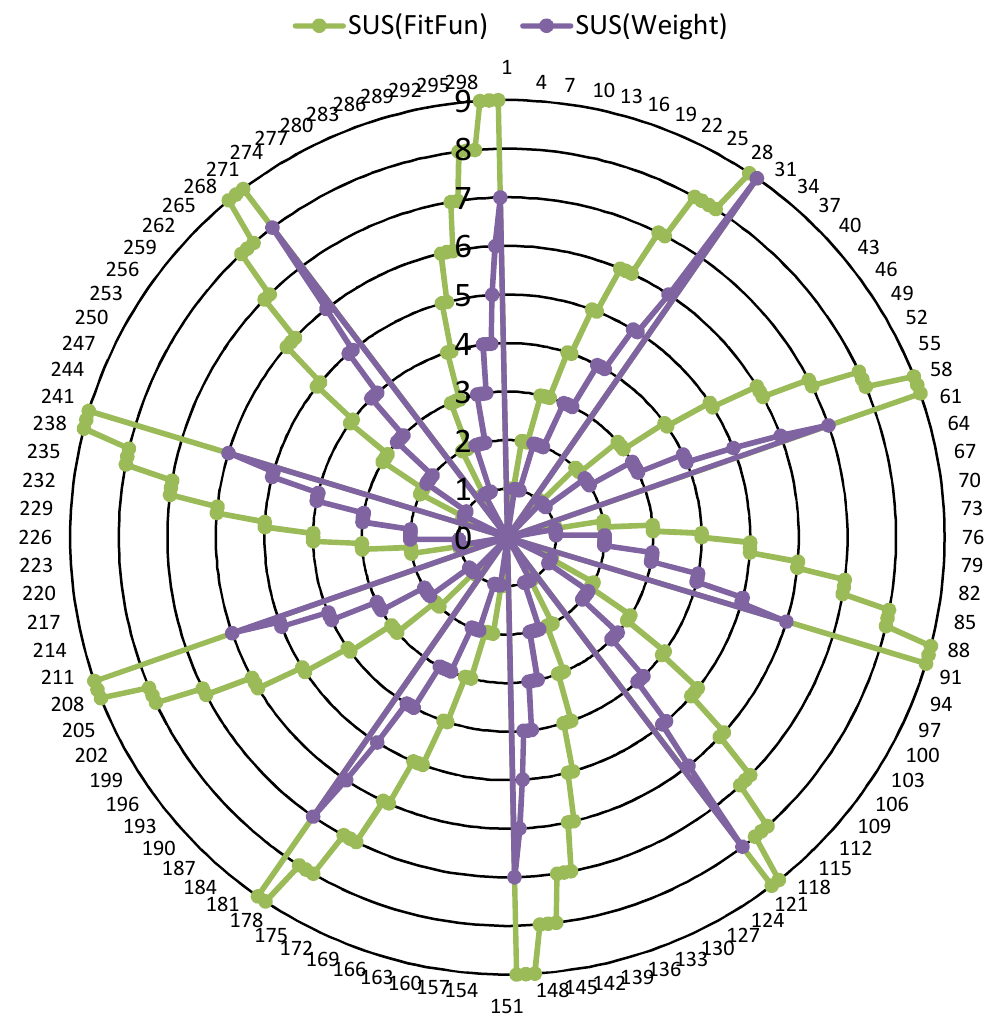} \includegraphics[width=1.5in, height = 1.5in]{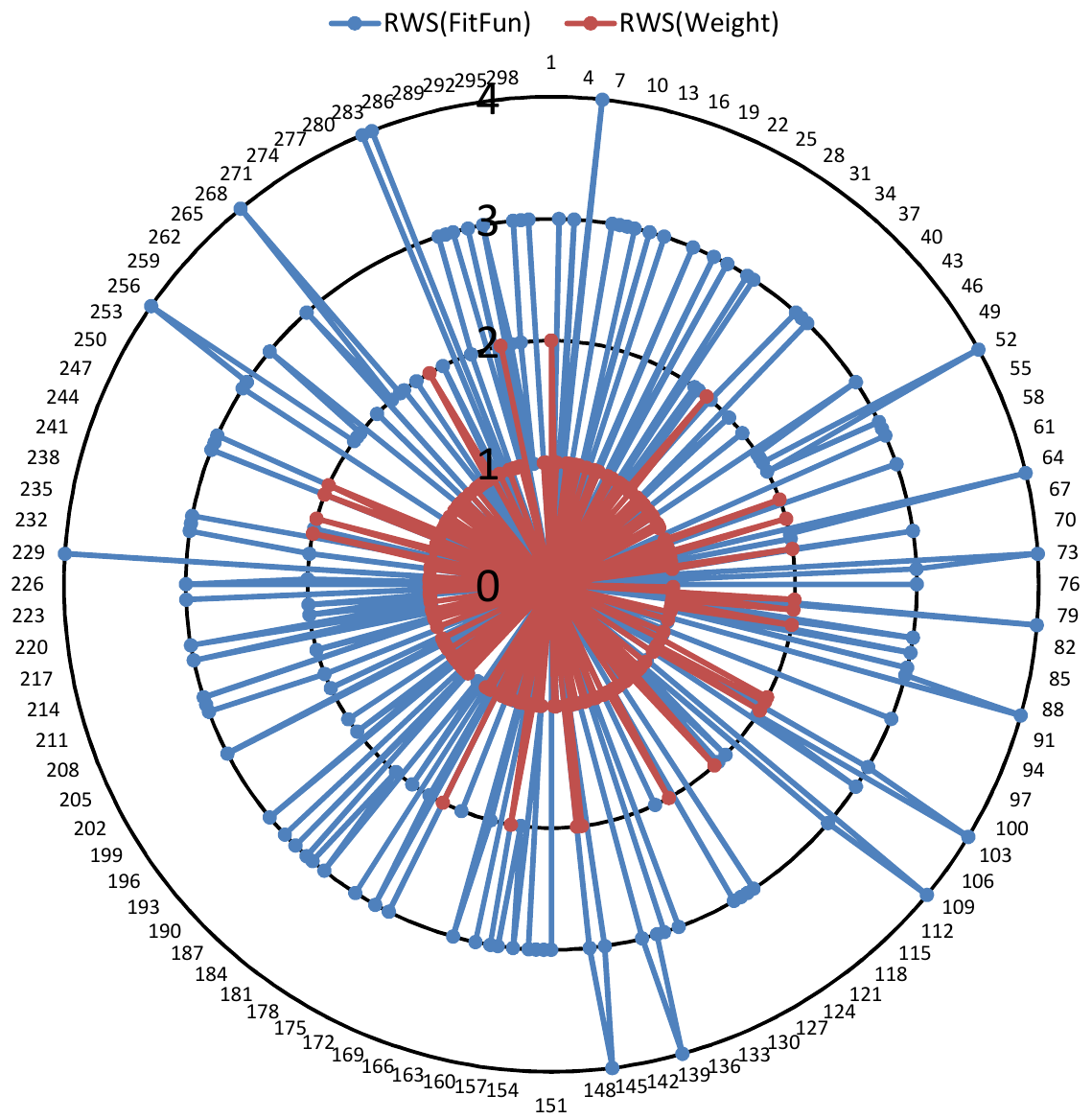}
\includegraphics[width=1.5in, height = 1.5in]{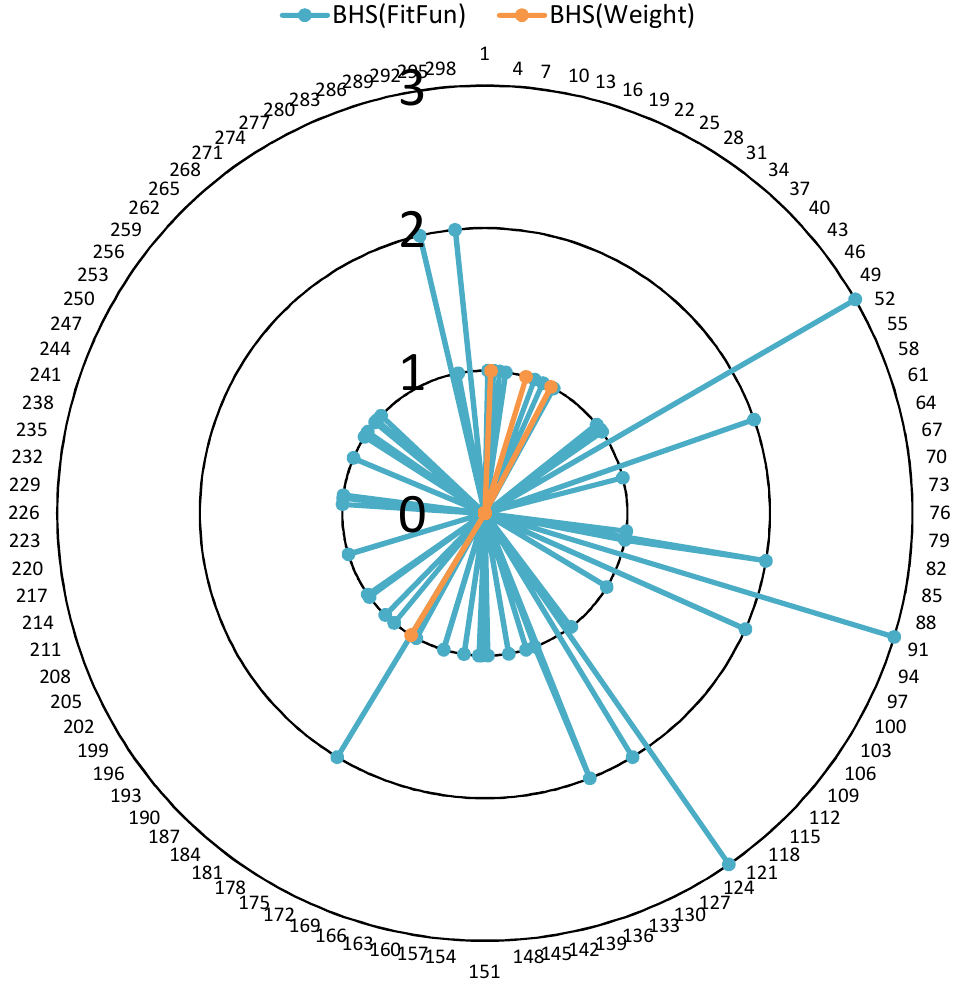} \includegraphics[width=1.5in, height = 1.5in]{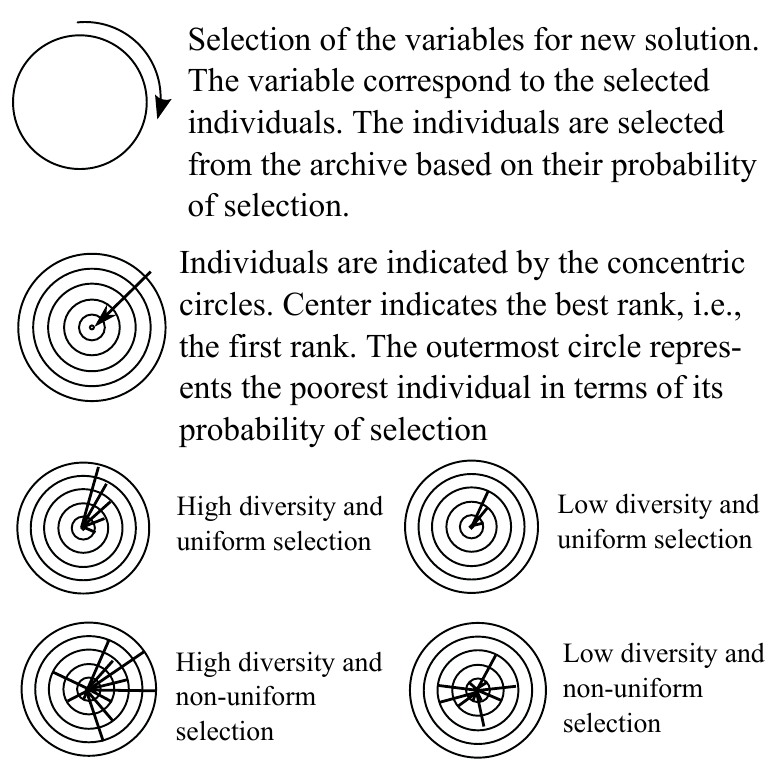}
\caption{Mapping of the individual selection according to their rank. The mapping indicate the span of rank coverage in selection and the distribution of selection. (a) top left - SUS( green - FitVal, purple - weight), (a) top right - RWS( indigo - FitVal, red - weight) and (a) bottom left - BHS( blue - FitVal, orange - weight)}
\label{fig:selInd}
\end{figure}

The six selection strategies namely, RWS(FitVal), RWS(Weight), SUS(FitVal), SUS(Weight), BHS(FitVal) and BHS(Weight) are represented by indigo, red, green, purple, blue and orange coloured lines. Examining Figures \ref{fig:selInd}(a) and \ref{fig:sel}, SUS(FitVal) and SUS(Weight) selection strategy selects the solutions uniformly in each of the new individual construction. The distribution of the selection (picking a solution) is uniform (from best to worst) throughout 300 selection. Therefore, the selection are repeat at a step of 30 (dimension). It may also be observed that the coverage of selection is distributed from the best to worst selection. The results provided in Table \ref{tab:sel_sse} indicates poor performance of SUS selection strategy. Therefore, an uniform selection with wider coverage of ranks happens to be poor selection strategy. In case of RWS(FitVal) and RWS(Weights) selection mapping provided in Figures \ref{fig:selInd}(b) and \ref{fig:sel} indicated a span of selection from rank 1 (center marked 0) to rank 5 (outermost circle marked 4) in case of function value and rank 1 to rank 3 in case of Weight. However, its selection is mostly distributed within the range of rank 1 to 4 in case of FitVal and rank 1 to 2 in case of Weight. It is worth mentioning that unlike SUS case, in RWS the selections is not uniform throughout 300 selections for construction of each of the 10 new solutions. The non uniform selection with coverage of selection to an adequate range of best to worst solution helps RWS selection strategy to achieve better performance over its competitor selection strategy. Similar to RWS, BHS selection also offer non uniform selection of individuals but on the contrary to the RWS its coverage of rank is mostly concentrated to fittest individual in the archive. From Figures \ref{fig:selInd}(c) and \ref{fig:sel}, it may be observed that the BHS selection is non-uniform but its selection spans upto rank 3 among the 10 individuals whereas the RWS spans upto rank 5. From  Figures \ref{fig:selInd}(a), \ref{fig:selInd}(b), \ref{fig:selInd}(c) and \ref{fig:sel}, it may observed that probability of selection computed based on the weights indicated in purple (in Figure \ref{fig:selInd}(a)), red (in Figure \ref{fig:selInd}(b)) and orange (in Figure \ref{fig:selInd}(c)) behaves similar to the probability of selection computed based on function value but, it tend to prefer selection towards best ranks. However, the results provide in Table \ref{tab:sel_sse} indicated the the preference of better rank in case of SUS offers better result than preferring each individuals in archive.     
\subsection{Distance Measure Metric}
\label{subsec:pa_dist}
After the selection of a solution from an archive, another crucial operation in ACO algorithm is sampling of the selected solution. To sample the selected solution, parameter $\mu$ and $\sigma$ need to be computed. As discussed in section \ref{sec:caco}, the parameter $\mu$ is the $i^{th}$ variable of the $j^{th}$ solution selected for sampling and the parameter $\sigma$ is computed as per expression (\ref{eq:sigma_comn}) where distance $D_i$ ($i \in [1, 10]$)  is a distance metric listed in Table \ref{tab:dist}. The distance computed between the selected solution (point) with other solution (points) in the solution archive is critical to the performance of ACO algorithm. To compute distance between two points, usually the Euclidean distance metric is used. In general for computing distance between points ($x_1$, $x_2$, $ \ldots $, $x_n$) and ($y_1$, $y_2$, $ \ldots $, $y_n$) Minkowski distance of order r is used. The Euclidean distance is a spacial case of Minkowski distance metric, where r = 2. An experimental result over all the distance metric mentioned in the Table \ref{tab:dist}. The experimental setup is as follows: $k = 10$, $n =2$, $m = 10$, $\xi = 0.5$ and $ \epsilon =$ 1000 iterations. It may be noted that for the experimentation purpose the RWS selection with probability selection based on function value is used. Examining table \ref{tab:dist}, it is found that the Squared Euclidean (D6) is performed better than all the other distance metric. However, it may also be noted that the performance of ACO decreases over the increasing order of 'r' of Minkowski metric.
\begin{table}[!t]
\caption{An experimental result over the distance measure metric indicated that superiority of squared euclidean distance over all other distance metrics}
\label{tab:dist}
\centering
\begin{tabular}{c|c|c|c}
\hline
\# & \multicolumn{2}{c|}{Distance Measure Metric} & Mean Fun. Value \\
     \cline{2-3}     
   & \multicolumn{1}{c|}{Expression} & \multicolumn{1}{c|}{Metric Name} & \\
\hline
 D1  & \multirow{6}{*}{$(\sum | x_i - y_i |^r )^{1/r}$} & Minkowsky (r = 0.5) & 28.792   \\
 D2  &                        & Manhattan (r = 1)   & 33.203  \\
 D3  &                        & Euclidean (r = 2)   & 44.578  \\
 D4  &                        & Minkowsky (r = 3)   & 45.211   \\
 D5  &                        & Minkowsky (r = 4)   & 51.909   \\
 D6  &                        & Minkowsky (r = 5)   & 53.702   \\
 \hline
 D7 &  $\sum (x_i - y_i )^2$ & Squared Euclidean    &  14.308  \\
\hline
 D8  &    $\max | x_i - y_i |$  & Chebychev         &  93.642  \\
\hline 
 D9  &   $ \frac{\sum | x_i - y_i |}{\sum x_i + y_i} $  & Bray Curtis &  98.983  \\
\hline
 D10  &   $\sum \frac{| x_i - y_i |}{| x_i| + |y_i |} $  & Canberra  &  103.742  \\
\hline
\end{tabular}
\end{table}

\subsection{Evaporation Rate}
\label{subsec:pa_xi}
\begin{figure}[!h]
\centering
\includegraphics[width=3in]{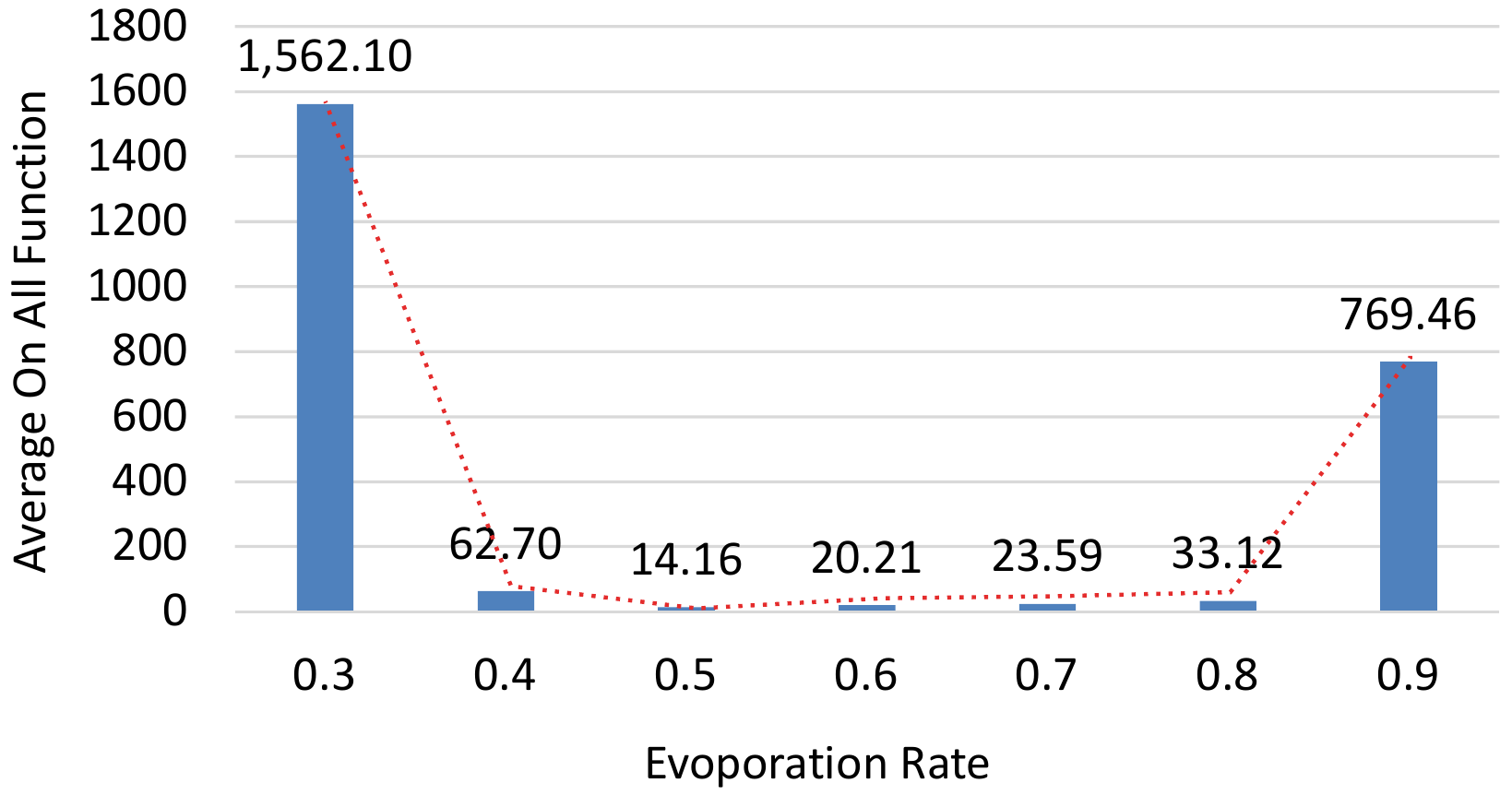}
\caption{An experimental results over different value of evaporation rate $\xi$. The results indicate a high sensitivity towards the parameter $\xi$.}
\label{fig:xi}
\end{figure}

The parameter evaporation rate ($\xi$) in ACO algorithm is treated as learning rate. The performance evaluation of ACO based on the evaporation rate with the following parameter combination $k = 10$, $n = 20$, $m = 10$, and $ \epsilon =$ 1000 iterations is illustrated in Figure \ref{fig:xi}. It may be noted that for the experimentation purpose the RWS selection with probability selection based on function value is used and the distance metric D6 is chosen for the computation of $\sigma$. In Figure \ref{fig:xi}, the values along the vertical axis represents the mean functions F1 to F10 (where each function value is averaged over twenty trails) listed in Table \ref{tab:testProblem} while the values along the horizontal axis represent evaporation rate $\xi$. The performance of ACO is evaluated by regulating the evaporation rate between 0.1 and 0.1. Investigating Figure \ref{fig:xi}, it may be observed that a valley shaped curve is formed. Initially, for $\xi = 0.3$, a high mean of function value is noted. While changing the value of $\xi$, a substantial improvement is being observed in the performance of the ACO. Hence, the performance of ACO is highly sensitive to the parameter $\xi$. It may be observed from Figure \ref{fig:xi}, that the increasing the value of $\xi$ enhance the performance of ACO. However, the performance of ACO slightly declined on further increasing the value of $\xi$ onward 0.5. A sudden high drop in performance if observed at the evaporation rate high sensitivity towards evaporation rate can be observed from the Figure \ref{fig:xi}.         

\subsection{Comparison with Other metaheuristics}
An experiment conducted for the comparison between the improvised ACO and other classical metaheuristic algorithm such as Particle Swarm Optimization (PSO) and Differential Evaluation. 

The parameter setting adopted for ACO is as follows: the population $k = 10$, $n = 20$, and $ \epsilon =$ 1000 iterations, evaporation rate $\xi$ is set to 0.5, selection method for classical ACO is BHS(Weight) and distance metric is D2 (Manhattan) whereas for ACO$^*$ (improvised parameters) selection method is RWS(FitVal) and the distance metric is D6 (Squared Euclidean) is considered. 

The PSO \cite{psoEberhart1995} is a population based metaheuristic algorithm inspired by foraging behaviour of swarm. A swarm is basically a population of several particles. The mechanism of PSO depends on the velocity and position update of a swarm. The  velocity in PSO is updated in order to update the position of the particles in a swarm. Therefore, the whole population moves towards an optimal solution. The influencing parameters are cognitive influence $C_1$, social influence $C_2$ are set to 2.0 and $Inertia Weight High$ is set to 1.0 and $Inertia Weight Low$ is set to 0.0. The other parameter population size set to 10 and  
$ \epsilon$ set to 1000 iterations.

The DE \cite{deStorn95,deStorn1997} inspired by natural evolutionary process is a popular metaheuristic algorithm for the optimization of continuous functions. The parameter of DE such as $weight factor$ is set to 0.7, the $corssover factor$ is set to 0.9 are the major performance controlling parameter. In present study the DE version $DE/rand-to-best/1/bin$ \cite{deQin2009} is used. The other parameter population size set to 10 and $ \epsilon$ set to 1000 iterations. Examining Table \ref{tab:comRes}, it may be concluded that at the present mentioned experimntal/ parameter setup, the improvised version of ACO outperform the classical metaheuristics. However, from the present paper and study and the availability of no free lunch theorem \cite{wolpert1997no}, it is clearly evident that the mentioned meta-heuristcs are subjected to parameter tuning. Hence a claim of superiority of the present improvised ACO is subject to its comparisons parameter tuning of the other mentioned metaheuristics.    
\begin{table}[t]
\centering
\caption{An experiment conducted for comparison between classical metaheuristics algorithms. The results indicate performance superiority of ACO algorithm over other algorithm mentioned in table. \tiny{\textbf{Note:} ACO indicates the original version of ACO, ACO indicates improvised ACO, PSO - Classical Particle Swarm Optimization and DE - Deferential Evolution.}}
\label{tab:comRes}
\fontsize{8}{10}
\selectfont
\setlength\extrarowheight{2pt}
\begin{tabular}{l|l|r|r|r|r}
\hline
	
\multicolumn{1}{c|}{Funtion}& \multicolumn{1}{c|}{Test}& \multicolumn{1}{c|}{ACO}  & \multicolumn{1}{c|}{ACO} &  \multicolumn{1}{c|}{PSO} & \multicolumn{1}{c}{DE}  \\
\hline
F1 & $ f(x^*) $ & 1.72 & 1.63 & 17.86 & 11.16 \\
  & var & 0.05 & 0.01 & 3.80 & 13.92 \\
F2 & $ f(x^*) $ & 0.69 & 0.02 & 7875.01 & 1610.96 \\
  & var & 0.02 & 0.00 & 1.45E+07 & 3402.52 \\
F3 & $ f(x^*) $ & 5.57 & 0.47 & 488.92 & 40.81 \\
  & var & 0.73 & 4.17 & 2.00E+05 & 200.54 \\
F4 & $ f(x^*) $ & 131.42 & 65.23 & 2.20E+05 & 2763.53 \\
  & var & 4501.12 & 6160.92 & 9.35E+09 & 37334.53 \\
F5 & $ f(x^*) $ & 127.56 & 32.24 & 81.27 & 22.27 \\
  & var & 308.44 & 618.88 & 569.12 & 56.08 \\
F6 & $ f(x^*) $ & 0.46 & 0.06 & 62.64 & 13.22 \\
  & var & 0.27 & 0.01 & 712.26 & 39.88 \\
F7 & $ f(x^*) $ & 4.93 & 12.72 & 458.32 & 44.71 \\
  & var & 1.60 & 294.76 & 20947.18 & 149.05 \\
F8 & $ f(x^*) $ & 11.68 & 1.05 & 36556.80 & 2162.60 \\
  & var & 324.92 & 0.03 & 4.18E+09 & 13945.71 \\
\hline
\end{tabular}
\end{table}
\section{Discussion and Conclusion}
\label{sec:con}
A comprehensive performance analysis of Ant Colony Optimization is offered in present study. The parameter such as selection strategy, distance measure metric and evaporation rate are put into meticulous tuning. The selection of a variable in construction of new solution. The assignment of the probability of selection to the individuals in the selection strategy influence the performance ACO. Analysing the results produce by the various selection strategy, it  may be conclude that the selection strategy, RWS together with the probability of selection computed based on the function value offer better result than its counterparts. The advantages with RWS strategy is due to its ability to maintain non uniformity in selection and prefering not only the best solution in a population of individuals. Rather than sticking to Manhattan distance metric only, it is interesting to test several available distance measure metric for computing average distance between the selected solution and all the other solutions. It is observed from the experiments that the distance metric Squared Euclidean offer better performance among the mentioned distance metric in present study. It may also observed from the analysis that the ACO is highly sensitive towards its parameter, pheromone evaporation rate which control the magnitude of the average distance between the selection solution (variable) to all the other solution (individuals) in the population. A comparison between classical metaheuristic indicated the dominance of the ACO algorithm in present experimental setup. However, as evident from the present study and the no free lunch theorem, the metaheuristic algorithms are subjected to parameter tuning. Therefore, claim of superiority of one metaheuristic algorithm over other will always be under scanner.      
\section*{Acknowledgment}
This work was supported by the IPROCOM Marie Curie initial training network, funded through the People Programme (Marie Curie Actions) of the European Union's Seventh Framework Programme FP7/2007-2013/ under REA grant agreement No. 316555.

\ifCLASSOPTIONcaptionsoff
  \newpage
\fi

\bibliographystyle{IEEEtran}
\bibliography{vkoISDA}
\end{document}